\documentclass[12pt]{article}

\usepackage{newtxtext,newtxmath}

\usepackage{graphicx}
\usepackage{booktabs} 
\usepackage[letterpaper,margin=1in]{geometry}
\usepackage{booktabs}
\usepackage{array}
\renewenvironment{abstract}
    {\centering \section*{Abstract}%
   \addcontentsline{toc}{section}{Abstract}%
   \begingroup\bfseries}
  {\endgroup}

\date{}

\makeatletter
\renewcommand{\fnum@figure}{\textbf{Figure \thefigure}}
\renewcommand{\fnum@table}{\textbf{Table \thetable}}
\makeatother

\usepackage{scicite}
\usepackage{makecell}
\usepackage{url}
\usepackage{pdfpages}
\usepackage{caption} 
\usepackage{comment}

\def\scititle{Ecological Legacies of Pre-Columbian Settlements Evident in Palm Clusters of Neotropical Mountain Forests}
\title{\bfseries \boldmath \scititle}

\author{
Sebastian Fajardo$^{1\ast\dagger}$,
    Sina Mohammadi$^{1\ast\dagger}$,
    Jonas Gregorio de Souza$^{2}$,
    César Ardila$^{3}$,\and
    Alan Tapscott$^{4}$,
    Shaddai Heidgen$^{4}$,
    Maria Isabel Mayorga Hernández$^{5}$,
    Sylvia Mota \and de Oliveira$^{6}$,
    Fernando Montejo$^{3}$,
    Marco Moderato$^{4}$,
    Vinicius Peripato$^{7}$,
    Katy Puche$^{3}$, \and
    Carlos Reina$^{3}$,
    Juan Carlos Vargas$^{8}$,
    Frank W. Takes$^{1}$,
    Marco Madella$^{4,9,10\ast}$\and
    \small$^{1}$ Leiden Institute of Advanced Computer Science (LIACS), Leiden University, Leiden, The Netherlands 2300 RA\and
    \small$^{2}$ Computational Social Sciences and Humanities Laboratory, Barcelona Supercomputing Center - \and \small Centro Nacional de Supercomputación (BSC-CNS). Barcelona, Spain 08034\and
    \small$^{3}$ Subdirección de Gestión del Patrimonio, Instituto Colombiano de Antropología e Historia, Bogotá, Colombia 111711\and
    \small$^{4}$ Department of Humanities, CASEs group, Universitat Pompeu Fabra, Barcelona, Spain 08023\and
    \small$^{5}$ Facultad de Artes, Universidad Nacional de Colombia, Bogotá, Colombia 111321\and
    \small$^{6}$ Naturalis Biodiversity Center, Leiden, The Netherlands, 2300 RA\and
    \small$^{7}$ Division of Earth Observation and Geoinformatics (DIOTG), General Coordination of Earth Sciences (CG-CT),\and \small National Institute for Space Research (INPE), São José dos Campos, SP, Brazil\and
    \small$^{8}$ Departmento de Antropología, Universidad del Magdalena, Santa Marta, Colombia 470004\and
    \small$^{9}$ ICREA, Barcelona, Spain 08010\and
    \small$^{10}$ GAES, The University of the Witwatersrand, Johannesburg, South Africa 2050\and
	\small$^\ast$Corresponding authors Sebastian Fajardo: s.d.fajardo.bernal@liacs.leidenuniv.nl;\and 
    \small Sina Mohammadi: s.mohammadi@liacs.leidenuniv.nl;\and \small Marco Madella: marco.madella@upf.edu\and
	\small$^\dagger$These authors contributed equally to this work.
}
\usepackage{lineno}  
\usepackage{listings} 

\begin{document} 

\maketitle

\begin{abstract}
Ancient populations inhabited and transformed neotropical forests, yet the spatial extent of their ecological influence remains underexplored at high resolution.
Here we present a deep learning and remote sensing based approach to estimate areas of pre-Columbian forest modification based on modern vegetation.
We apply this method to high-resolution satellite imagery from the Sierra Nevada de Santa Marta, Colombia, as a demonstration of a scalable approach, to evaluate palm tree distributions in relation to archaeological infrastructure.
Our findings document a non-random spatial association between archaeological infrastructure and contemporary palm concentrations. Palms were significantly more abundant near archaeological sites with large infrastructure investment. 
The extent of the largest palm cluster indicates that ancient human-managed areas linked to major infrastructure sites may be up to two orders of magnitude bigger than indicated by current archaeological evidence alone. These patterns are consistent with the hypothesis that past human activity may have influenced local palm abundance and
potentially reduced the logistical costs of establishing infrastructure-heavy settlements in less accessible locations. More broadly, our results highlight the utility of palm landscape distributions as an interpretable signal within environmental and multispectral datasets for constraining predictive models of archaeological site locations.
\end{abstract}


\section*{Introduction}
\noindent
The activities of ancient populations transformed the composition of forests in many neotropical regions. Ancient human management included the cultivation, tending, and dispersal of tree species for food and technological materials \cite{levis_persistent_2017,coelho_eighty-four_2021}, as well as the use of fire for forest clearance and enriching nutrient soil composition \cite{leal_human_made_2019,maezumi_fire_human_climate_2023}. Forest tree inventory data, reported archaeological sites, and classic machine learning models have been used to examine the relationship between plant species and their association with past human activities in regions with tens of thousands of square kilometres \cite{levis_persistent_2017,peripato_more_2023,sales_potential_2022}. These studies show that the impacts of past activities on modern forest structures near the location of ancient settlements are long-lasting, to the extent that they may influence our current estimates of forest composition in tropical forests, such as the Amazon \cite{mcmichael_comment_2017, mcmichael_ancient_2017}, the eastern flank of the tropical Andes \cite{sales_potential_2022}, and the neotropical forests of Central America \cite{brokaw2025ancient,harvey2019legacy,hightower2014quantifying}.

Archaeological research documents frequent occupation of montane environments in tropical Andean regions \cite{sales_potential_2022}, including Colombia \cite{langebaek2010cuantos, botiva1989colombia}. The persistent fog and light limitations \cite{Letts_Mulligan_2005, Fahey_Sherman_Tanner_2016}, low soil fertility \cite{igac2009estudio,Fahey_Sherman_Tanner_2016}, and abrupt terrain \cite{Crausbay_Martin_2016} create conditions in these montane forests that required adaptive land-use and resource-management practices rather than conventional cultivation practices observed in lowland or intensive agrarian systems. While the distribution of tree species exploited for food is relatively well documented \cite{iriarte2020106582,Clement2021,Piperno2011}, the spatial relationships between plants that could potentially be used in the past for raw materials and ancient settlements of different sizes and spatial arrangements remain poorly understood. 

Efforts to estimate the effect of ancient human activities in modern neotropical forest composition, using archaeological sites and forest tree inventories, are constrained by limited data on the variability of pre-Columbian settlement patterns and on occurrences of tree species across the areas once occupied by these communities. Both limitations result in coarse resolution estimates, making it difficult to accurately assess the long-term effects of various past human activities on local forest composition \cite{clement_domestication_2015}. The challenge of detecting variability in settlement patterns and its relation with the distribution of tree species that could have been used in the past is especially evident in montane forests. Understanding how ancient human activities influenced neotropical montane forests requires analyses that cover spatially continuous areas large enough to capture both ecological and archaeological variability. Such landscape-scale perspectives are important for integrating human legacies into ecological models and for interpreting modern forest composition in a long-term context. However, traditional field-based approaches are difficult to apply across steep, densely forested montane terrain. Automated remote-sensing approaches, such as the method used here, provide a scalable way to examine relationships between forest composition and archaeological evidence across large and otherwise inaccessible regions. There is no uncertainty about whether montane regions were occupied---archaeological evidence clearly demonstrates sustained pre-Columbian presence \cite{sales_potential_2022,langebaek2010cuantos,botiva1989colombia}---but rather we need to measure how localised human activities may have shaped forest composition and structure across spatially continuous landscapes.

In this study, we propose a method called Past Areas of Bio-cultural Activity and Management (PABAM), an automated remote-sensing framework that integrates deep-learning–based palm detection with archaeological legacy data to examine potential human ecological legacies. This approach is designed to address the gap outlined above by linking continuous vegetation signals to spatially explicit archaeological data. Here, archaeological legacy data refers to information derived from historical maps, published archaeological reports, and previous surveys documenting site locations. By using these data sources, PABAM enables us to examine spatial relationships between archaeological sites and palm species with star-shaped crowns--those whose leaves radiate outward from the top of the trunk like the points of a star shape--associated with multi-purpose exploitation by indigenous populations\cite{Paniagua-Zambrana2020, Moraes2015}. 

We tested the method in the montane forests region of the Sierra Nevada de Santa Marta (SNSM) in Colombia. By combining deep learning to identify the locations of individual palms based on the star-shaped crown, alongside archaeological legacy data, we aimed to understand the association between the presence of these plants and the archaeological sites. While the PABAM method is tested in the SNSM, it is designed to be broadly applicable to other neotropical, and potentially global, regions with similar vegetation and archaeological complexity. This integrated approach provides deeper insights into the formation of cultural landscapes in neotropical forests. The PABAM framework does not aim to provide a definitive and standalone prediction of archaeological sites. Instead, it offers a useful supplementary signal that when combined with environmental covariates (e.g. soils, topography, microclimate) and multispectral or hyperspectral remote-sensing data, can help to prioritize landscape sectors for detailed, site-level proxy investigations. 

We hypothesize that present-day palm concentrations non-randomly co-occur with areas of intensive past human activity. To test our hypothesis, we analysed palm concentrations around archaeological sites using an Inverse Distance Weighting (IDW) score and conducted an elevation analysis comparing palm elevation near sites to regional baselines. The regional baselines refer to the elevational distribution of palms, specifically \textit{Dictyocaryum}, as documented in the Global Biodiversity Information Facility (GBIF), which serves here as an observation-based reference. These analyses allowed us to assess whether palm presence is related to archaeological site locations and palm elevation patterns indicative of ancient human management.

Our main contributions are as follows. 1) A method to analyse the relation between tropical vegetation and past human activity. We propose a multi-component machine learning method to investigate the spatial relationship between vegetation patterns and archaeological sites. 2) Release of a manually annotated palm tree dataset. We are releasing a high-quality palm tree dataset covering 69.5 km$^2$, manually labelled by human annotators (Figure \ref{fig:studyArea}). Researchers can use our annotated dataset as a source data to develop and test methods or adapt models to new regions. 3) Release of ground-surveyed archaeological site locations. We are also releasing the locations of archaeological sites in the area of study, along with associated levels of certainty, verified through ground surveys and reviews of legacy data. This dataset can support future studies exploring the relationship between the presence of archaeological sites, vegetation patterns, and past human activity.

\section*{Results}

\subsection*{Palm Tree Detection Model Performance}
We evaluated the performance of the YOLOv10 object detection model developed to identify star-shaped palm crown patterns as a morphological feature, rather than individual palm species, on the test dataset using different confidence thresholds of 0.2, 0.3, and 0.4. Precision and recall metrics are reported in Figure \ref{fig:method}. In our application, minimizing false positives was more important than maximizing recall because high false positive rates in palm tree detection could produce artificial clusters and misleading density estimates. Since clusters should reflect true patterns with high precision, we prioritized precision over recall. Therefore, a confidence threshold of \(\geq 0.4\) was selected to reduce false positives. The final palm tree bounding box predictions for the entire study area, based on the \(\geq 0.4\) threshold, are shown in Figure \ref{fig:method}.

\begin{figure*}[!t]
  \centering
  \includegraphics[scale=1]{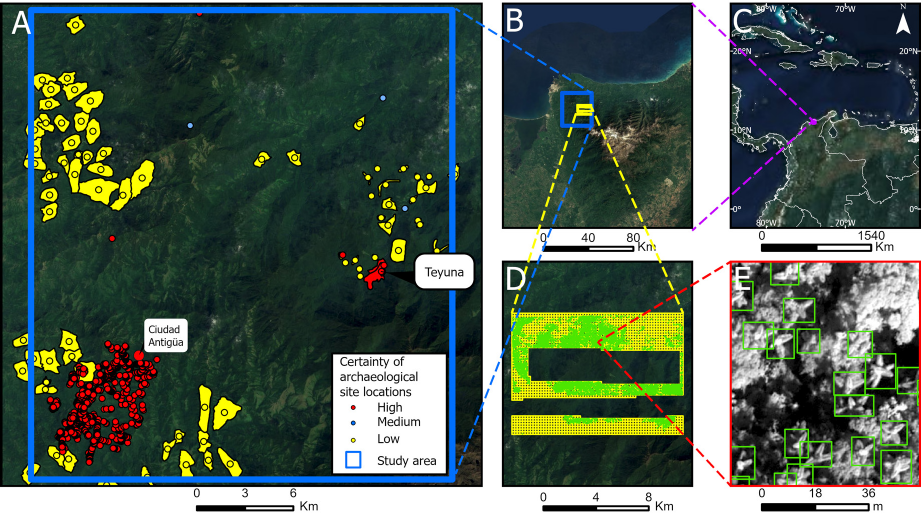}
  \caption{Study area at different resolutions. Regional (A and D) and Global (B and C) context of the study area. Archaeological site locations with varying levels of location certainty (A). Annotated 200×200 m grid in yellow and palm tree labels in green (D). Detailed view of annotated palm trees on high-resolution imagery (E). Panel E includes material © CNES (2024), Distribution Airbus DS. All other panels include Sentinel-2 cloudless - https://s2maps.eu by EOX IT Services GmbH (Contains modified Copernicus Sentinel data 2016 \& 2017).}
  \label{fig:studyArea}
\end{figure*}

\subsection*{Palm Distribution and Archaeological Sites}

\begin{figure*}[!t]
  \centering
  \includegraphics[scale=0.5]{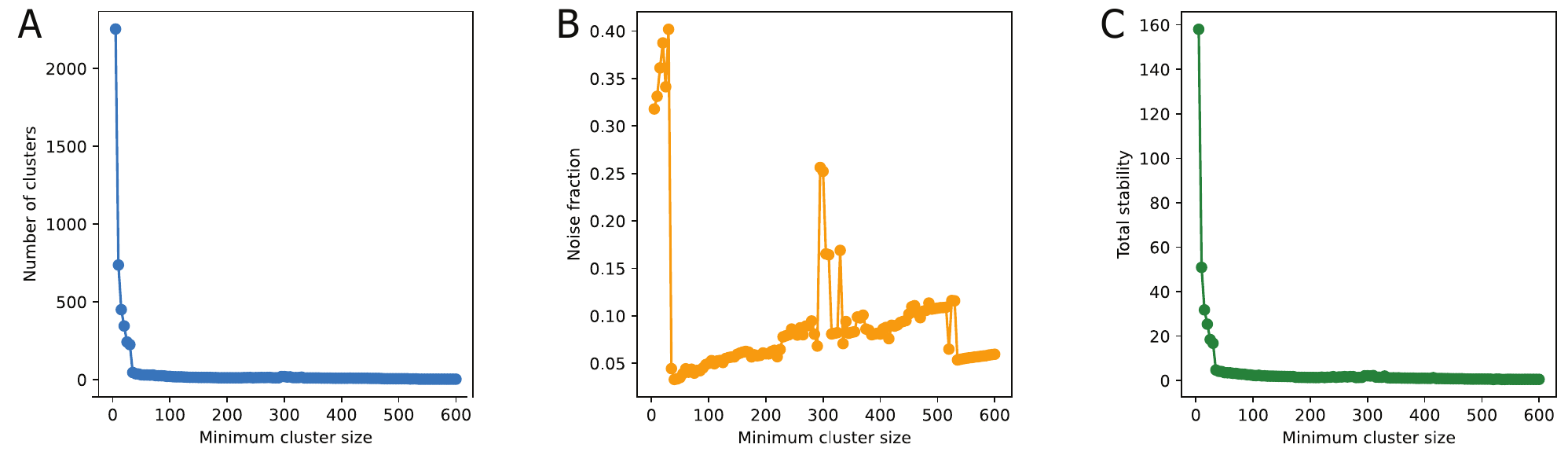}
  \caption{HDBSCAN parameter sweep showing the effects of min\_cluster\_size (5–600) on cluster number (A), noise fraction (B), and total cluster stability (C). Low thresholds produce numerous unstable clusters, whereas values $\ge$100 yield a stable set of robust palm clusters. The selected parameter value balances stability and noise reduction and supports the robustness of subsequent spatial analyses.}
  \label{fig:HDBSCAN_parameter_sweep}
\end{figure*}

To evaluate the relationship between palm tree distribution and past human presence, we conducted a spatial clustering analysis of palm trees across the study area. Because clustering outcomes can be sensitive to parameter choice, we systematically explored the effects of varying the minimum cluster size to assess the robustness of detected spatial groupings. The parameter sweep (Figure \ref{fig:HDBSCAN_parameter_sweep}) shows a non-linear relationship between minimum cluster size and the number of clusters. With a minimum cluster size of five, more than 2000 small clusters were detected (Figure \ref{fig:HDBSCAN_parameter_sweep}A). Around 100, the number of clusters stabilized, indicating more robust groupings. Total cluster stability (Figure \ref{fig:HDBSCAN_parameter_sweep}C) initially declined sharply before stabilizing. The noise fraction (Figure \ref{fig:HDBSCAN_parameter_sweep}B) increased gradually as more points were excluded from larger clusters. A minimum cluster size of 100 produced a number of stable clusters with high persistence and a noise fraction of approximately 5\%, indicating that most data points were retained while filtering out low-density groupings. Beyond this value, stability changed little, suggesting that remaining clusters persist across distance scales. Therefore, a conservative minimum cluster size of 100 was selected for the final analysis. 

\begin{figure*}[!t]
  \centering
\captionsetup{font=footnotesize}

  \includegraphics[scale=0.80]{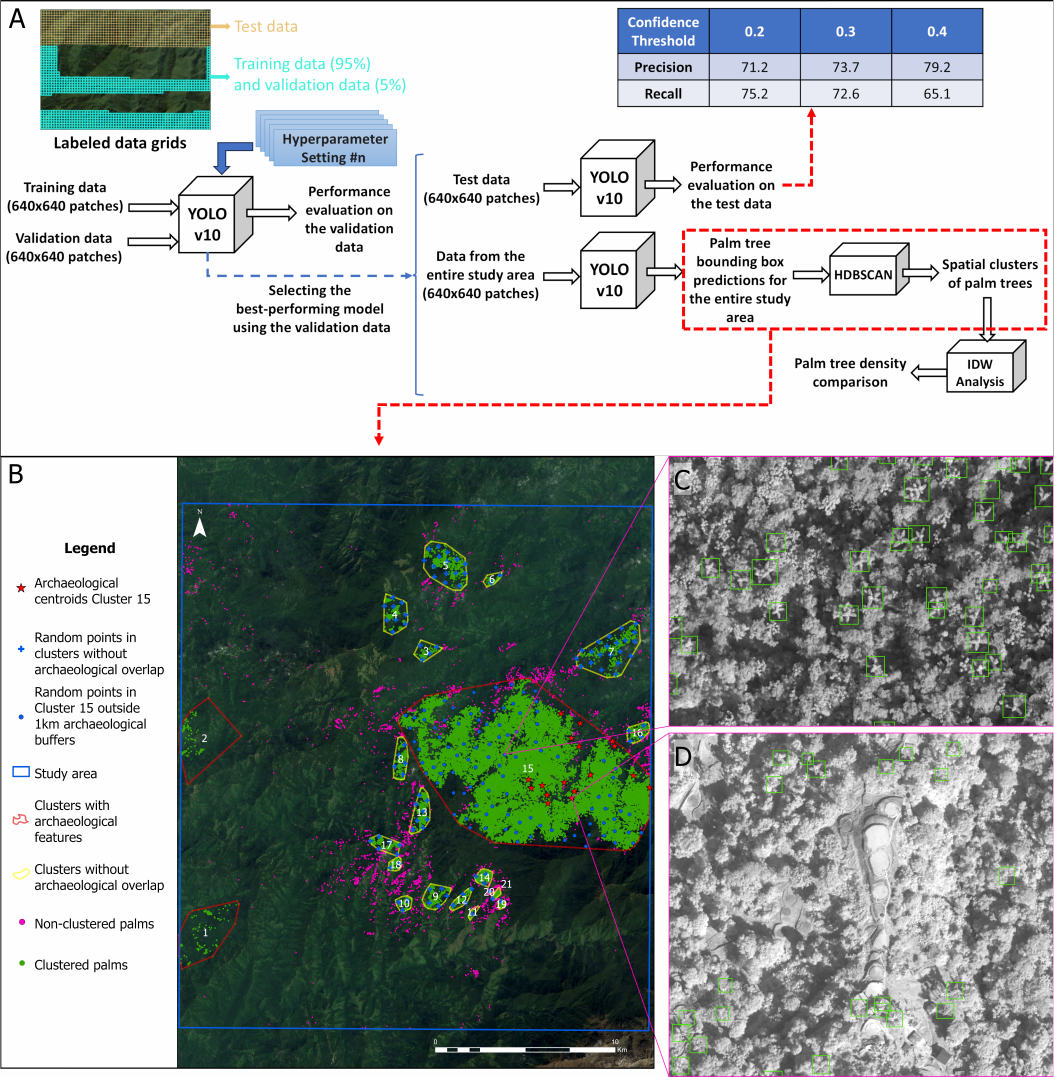}
  \caption{PABAM workflow diagram and palm tree detection and spatial clustering results. The PABAM workflow (A) illustrates the selection of the best-performing model based on validation data. This model was then applied to the test set for performance evaluation and to the entire study area for palm detection. Detected palm trees were subsequently clustered using the HDBSCAN algorithm, and an inverse distance weighting (IDW) analysis was performed to estimate spatial variations in palm density. The bottom-left panel (B) shows the identified palm clusters with their corresponding cluster IDs, distinguishing unclustered palms (magenta points) from clustered palms (green points), including clusters that spatially co-occur with archaeological sites (notably Clusters 15, 1, and 2) as well as clusters without associated archaeological features. The top-right panel (C) illustrates areas within palm Cluster 15 located more than 1 km from any known archaeological site. The bottom-right panel (D) shows detected palms near the center of Teyuna.   Panels  C and D include material © CNES (2024), Distribution Airbus DS. Panel B includes Sentinel-2 cloudless - https://s2maps.eu by EOX IT Services GmbH (contains modified Copernicus Sentinel data 2016 and 2017).}
  \label{fig:method}
\end{figure*}

\begin{figure*}[!t]
  \centering
  \includegraphics[scale=0.95]{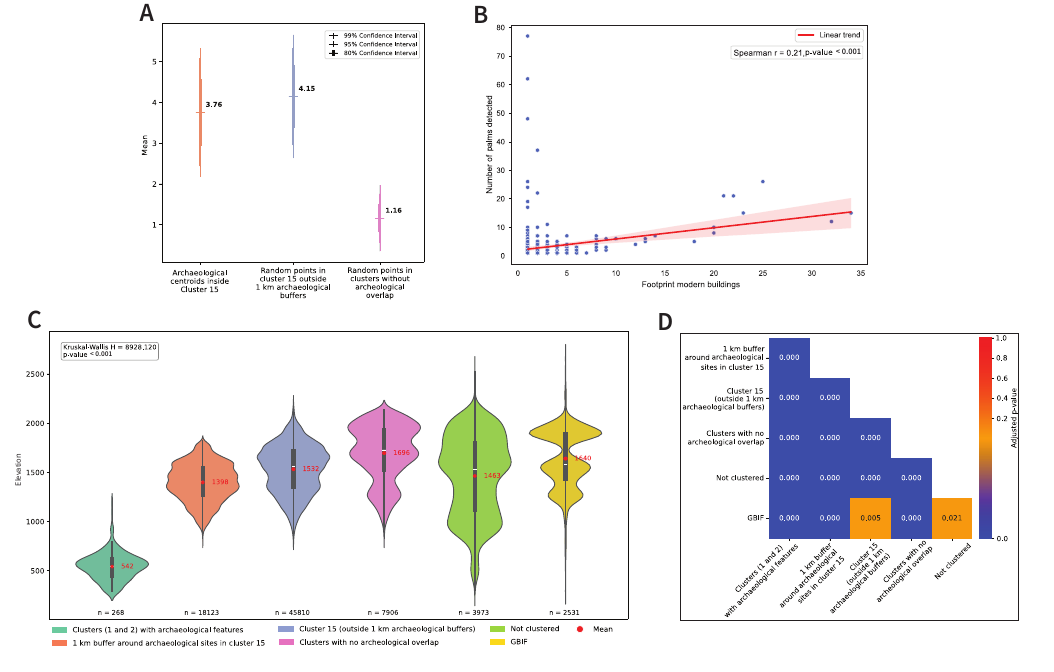}
  \caption{Human activity and palm distribution. (A) Bootstrapped confidence intervals for observed means (Bootstrap samples = 1000; sample size = 17). The subsets shown are IDW means for archaeological centroids inside Cluster 15 (left), 100 random points in Cluster 15 outside 1 km archaeological buffers (centre), and 100 random points in clusters without overlap with archaeological features (right). Means are shown as text labels to the right of each bullet graph. (B) Spearman correlation between modern buildings \cite{microsoft2023global} and palm detections in each 200x200 m grid inside the study area. (C) Elevation distributions across subsets of detected palms and the GBIF baseline for the distribution of \textit{Dictyocaryum} in South America. Violin plots display Gaussian kernel density estimates for each subset, with overlaid box plots indicating the median (white line) and mean (red dot). (D) The heatmap presents pairwise Dunn test results between elevation distributions of the subsets and the baseline in figure \ref{fig:footprint_and_palms}C following a significant Kruskal-Wallis test.}
  \label{fig:footprint_and_palms}
\end{figure*}

Clusters 1, 2, and 15, which are located in the southwest, west, and east portions of the study area, respectively (Figure \ref{fig:method}), exhibited spatial overlap with known archaeological features. Notably, the largest archaeological site (Teyuna) is located within the largest cluster of palms (Cluster 15), spanning approximately 100 km$^2$ across both well-lit ridges and steep shaded slopes in the eastern part of the study area (Figure \ref{fig:method}). This cluster also contains 17 archaeological zones recorded as points or polygons. We calculated the centroids of the sites inside Cluster 15 to serve as reference points for spatial analysis. This resulted in 17 reference points used in our study. Archaeological sites within the largest palm cluster are located in areas with significantly higher IDW values than random locations within palm clusters lacking archaeological features (Figure \ref{fig:footprint_and_palms}A). A comparison of palm tree densities between 100 randomly selected locations in Cluster 15 (excluding a 1 km buffer around archaeological sites) and the 17 archaeological locations within the same cluster shows that the mean densities overlap within the 80\% bootstrapped confidence interval. This overlap suggests that palm density does not differ significantly between archaeological and non-archaeological locations in this cluster (Figure \ref{fig:footprint_and_palms}A).

Beyond the localised analysis of archaeological centroids near Teyuna, we expanded the IDW analysis to include four additional archaeological centroids situated in geographically distinct regions of the study area. These centroids were spatially independent from both the Teyuna-associated centroids and the control points used in the initial analysis. One of these centroids, in the southwest of the study area corresponds to El Congo-Ciudad Antigüa \cite{Rodriguez_et_al_2023}, a previously known archaeological site. The remaining two centroids were estimated based on nearby archaeological sites reported in legacy data. The IDW values for the centroids in the northwest, southwest (Ciudad Antigüa), and south-centre regions were 0.005, 0.002, and 0, respectively. In contrast, the Teyuna centroid yielded an IDW value of 3.75. These results suggest that palm tree density around Teyuna is substantially greater than in other possible archaeologically central sites in the study area that are currently located in highly intervened landscapes. The intensity of contemporary land-use in those areas likely alters or masks palm distribution patterns, limiting direct comparability.

The spatial distribution of palm detections inside each cluster was characterised using Ripley's K-function and its linearised transformation, the L-function. Two independent analytical protocols were applied: a multi-group analysis covering all HDBSCAN clusters (except Cluster 15) and a dedicated analysis of Cluster 15, which required a distinct methodological treatment due to its exceptional spatial extent (see Materials and Methods). For the multi-group analysis, the dominant spatial pattern across both groups was aggregation. Group A showed a more mixed pattern, with one cluster classified as clustered (Cluster 2) and one as regular (Cluster 1), though the very small sample size of this group precludes any meaningful generalisation. In Group B, 16 of 18 clusters (89\%) were classified as clustered and two (Clusters 4 and 20) were classified as regular or overdispersed.  Cluster 20 showed an L(r) $-$ r curve that oscillates around zero across all evaluated distances, alternately crossing above and below the Complete Spatial Randomness (CSR) expectation without any sustained directional signal, and remaining largely within the 95\% simulation envelope. This pattern suggests that palm detections in this cluster show no meaningful tendency toward either aggregation or regularity at any of the spatial scales examined.  Cluster 1 and Cluster 4 were also classified as regular; however, visual inspection of their L-function curve suggests this classification may be a methodological artefact rather than a genuine spatial signal.  The observed L(r) $-$ r curve of Cluster 1 declines immediately below zero and remains strongly negative across all evaluated distances (Supplementary Material, Figure S1). Rather than indicating genuine regularity, this pattern reflects a density-mismatch artefact:  the CSR envelope inflates to values exceeding 800 by r = 800 m, rendering the null expectation entirely disconnected from the scale at which the observed point pattern operates. Similarly, the observed L(r) $-$ r curve of Cluster 4 rises to positive values peaking at approximately 20 at r $\simeq$ 100 m, a pattern indicative of clustering at fine to intermediate scales, but remains entirely below the upper CSR envelope across all evaluated distances (Supplementary Material, Figure S1). The CSR envelope itself expands steeply to values exceeding 150 by r = 300 m, reflecting an inflated null expectation driven by a low point density. Under these conditions, the Monte Carlo simulations generate point patterns far denser at larger distances than the observed data, systematically elevating the envelope and preventing the observed clustering signal from reaching significance. The regular classification for Clusters 1, 4, and 20 should therefore be interpreted with caution, and the result would benefit from reanalysis using a smaller study window or a subcell approach to obtain a more appropriate spatial null model. 

Mann-Whitney U tests revealed no statistically significant differences between groups for any of the eight clustering metrics (Supplementary Material, Table S3 and Table S4) evaluated (all p-value $>$ 0.15). The largest effect size observed was for pct\_r\_sig and pct\_above\_CSR\_hi (r = 0.639, p-value = 0.159), which approached but did not reach significance, suggesting a possible tendency toward greater clustering intensity in Group B relative to Group A. All other metrics showed negligible to small effect sizes (r $<$ 0.28, all p-value $>$ 0.58). These non-significant results must be interpreted with considerable caution with only two observations in Group A.

Given the exceptional size of Cluster 15 and the distinct analytical protocol applied to it (see Materials and Methods), its results are reported as a standalone characterisation and are not directly comparable with the metrics derived for the other clusters. The analysis provides clear evidence of significant spatial aggregation at fine to intermediate spatial scales. The observed L(r) $-$ r curve exceeded the upper CSR envelope across 96\% of evaluated distance bins (r\_sig\_first = 20 m, r\_sig\_last = 500 m), indicating that clustering was statistically significant across virtually the entire range of scales examined. The curve rose steeply from approximately 20 m, reaching L\_max = 31.3 at r = 500 m, with an L\_integral of 11174 reflecting strong and sustained aggregation. 

A consistent short-distance depression of L(r) $-$ r below the CSR expectation is visible at very small r (typically r $<$ 15–20 m) across almost all clusters analysed, regardless of archaeological association. This systematic inhibition zone at near-neighbour scales most likely reflects a minimum inter-detection spacing imposed by the algorithm rather than a genuine ecological signal.

\subsection*{Modern Buildings and Palm Tree Distribution}
The Spearman correlation assessing the relationship between the number of modern human buildings and palm tree counts within 200 × 200 m grid cells across the study area (Figure \ref{fig:footprint_and_palms}B) was low (\( \rho = 0.21\), p-value \(< 0.001\)), indicating a weak association between contemporary built infrastructure and palm tree distribution. This analysis was conducted to evaluate whether the palm clustering patterns observed near archaeological sites could be explained by modern human activities. The weak correlation suggests that present-day buildings do not account for the observed palm density patterns. This result is largely driven by small, nucleated modern settlements and low-density, dispersed farm populations concentrated at lower elevations on the western side of the study area, spatially separated from the areas containing the largest palm clusters.

\subsection*{Elevation Analysis}

The centroids of archaeological locations within Cluster 15 span elevations ranging from 819 to 1570 m, with a mean elevation of 1113 m. Figure \ref{fig:footprint_and_palms}C shows that palm trees located within a 1 km radius of archaeological sites in Cluster 15 tend to occur at lower elevations compared to those outside this buffer zone but within Cluster 15. Furthermore, both elevation distributions---palms near archaeological sites and those outside the buffer zone in Cluster 15---show more frequently values at lower elevations than those of other palm clusters without archaeological overlap. In addition, all median elevation values of clustered palms that overlap with archaeological sites are below the GBIF baseline median. In contrast, only the clusters of palms without archaeological overlap exhibit median elevation values higher than that of the GBIF baseline. 
Although all detected palm subsets fall within the GBIF elevation envelope---indicating that site-associated palms do not occupy environmentally novel ranges---there are differences in distributional patterns within this shared range. The contrast in the distribution shapes (Figure \ref{fig:footprint_and_palms}C) reveals that archaeological site-associated palms tend to occur at lower elevations compared to palms without archaeological association. Our interpretation therefore emphasizes within-range distributional shifts rather than categorical range separation as a possible signal of archaeological association.

\section*{Discussion}

\subsection*{Ecological Implications of Palm Distribution and Elevational Patterns}

Previous studies typically examined broad geographical regions using sparse sampling \cite{levis_persistent_2017,sales_potential_2022}, making it difficult to capture finer-scale patterns of human-environment interaction. Our fine-scale detection analysis across a contiguous, yet regionally extensive, area reveals a spatial co-occurrence between the pre-Columbian populations and the largest cluster of palm trees in the neotropical forest of the study area. This co-occurrence is not consistent across all sites but is particularly pronounced around Teyuna, where pre-Columbian populations invested significantly in infrastructure development. Teyuna is characterized by substantial engineered features, including terraced platforms and stone retaining walls, rammed-earth foundations, paved paths and stairways between Teyuna and other settlements, formal drainage systems, and tightly clustered domestic buildings. Together, these features are indicative of major infrastructural investments \cite{Giraldo2010,groot1985arqueologia}. Settlement data from the southwestern part of our study area indicate that most archaeological sites are relatively small (\(\leq\) 1 ha), with a maximum surface of around 5 ha \cite{vargas2022arqueologia,Soto2025}. This surface is considerably smaller than the main site of Teyuna (approximately 33 ha). The area around Teyuna contains a significantly higher number of individuals from the detected plant taxa than would be expected under a random distribution model. This non-random pattern is consistent with a possible co-occurrence with past human settlements. Alternative explanations, such as preferential settlement in naturally palm-rich areas or shared environmental factors, cannot be ruled out with our current data. Given that the estimated ancient human management area at the Teyuna site alone is around 100 ha \cite{Rodriguez_et_al_2023}, incorporating the large surrounding cluster of detected palms suggests a much broader area of influence. The observed statistical overlap in palm density between archaeological centroids and random points within Cluster 15 (Figure \ref{fig:footprint_and_palms}A) indicates that high palm concentrations are not merely localised to site centres. Instead, this uniform density suggests the entire $100\text{ km}^2$ cluster represents a possible broad zone of ancient management. If the detected pattern reflects ancient activities around Teyuna---including paths, cultivation, resource collection, and general forest management---then the ancient human footprint in the area could be up to two orders of magnitude larger than what archaeological evidence alone indicates. Similar dynamics have been observed in many parts of the Amazon, where archaeological earthworks coincide with the hyperdominance of useful species \cite{peripato_more_2023,levis_persistent_2017} and with the management of forest composition \cite{Maezumi2022IndigenousLandUseBolivianAmazon, Furquim2023ConstructedBiodiversity}, in the neotropical forests of Central America \cite{brokaw2025ancient, Harvey2019LegacyPreColumbianFire, Hightower2014AncientMayaLandUseLegacy}, and in the old world tropics \cite{Summerhayes2017ForestExploitationNorthernSahul,Dufraisse2022TamingTrees}.

Teyuna and adjacent sites located within the largest identified palm cluster exhibit significantly higher palm densities compared to randomly selected points inside clusters without evidence of prehispanic occupation. This spatial association could arise through several mechanisms. Pre-Columbian activities may have managed the forest and opened the canopy, creating favourable conditions for palms. Yet palms may also function as disturbance-tolerant or early- to mid-successional taxa that increase after natural canopy openings, landslides, or windthrow, meaning palm enrichment could occur through normal forest dynamics without human involvement. In both cases, palm re-colonization of secondary forests can extend over decades or longer, and some species remain associated with disturbed or undisturbed forest conditions even after more than 100 years of succession, reflecting both re-colonization lags and persistent vegetation structure differences that may favour palms in secondary forests \cite{Eiserhardt2011determinantspalms}. In addition, the typically short dispersal distances in palms combined with the low density of other species probably contributed to the higher local density. In the case of \textit{Dictyocaryum lamarckianum}, its stilt roots provide increased mechanical support on steep hillsides, offering an advantage in both naturally disturbed and potentially human-modified landscapes. If palms were already abundant, their technological utility---offering materials for construction, containers, thatch, and edible parts like palm hearts \cite{Paniagua-Zambrana2020, Moraes2015}---may have influenced settlement location decisions, creating the observed spatial co-occurrence. Given Teyuna's historical prominence as a major pre-Columbian centre \cite{Rodriguez_et_al_2023,Giraldo2010}, it is plausible that anthropogenic influences, such as the planting, cultivation, or preferential retention of palms, were present in this region. These practices may have constituted a form of low-intensity management or stewardship of plant communities \cite{coelho_eighty-four_2021}. Our results show that the spatial association between palm clusters and archaeological sites is not uniform across the landscape. Instead, it is spatially variable, with the strongest co-occurrence observed near the major centre of Teyuna. Further research is needed to quantify the relative contribution of human management legacy, site selection in palm-rich areas, differential preservation and detection of archaeological features, and to establish the causal relationships among these possible mechanisms. 

Elevation analysis reveals that palms associated with archaeological sites tend to occur at lower elevations than both the regional GBIF baseline and palms in non-site clusters. In the SNSM lower elevations are correlated with increased taxa competition for solar radiation due to the hilly terrain and persistently cloudy conditions. If Pre-Columbian populations actively managed these areas, they may have created micro-environmental conditions such as increased light availability, modified hydrology, or reduced interspecific competition that favoured palm survival at lower elevations. Such micro-environmental modifications are not without precedent: analogous processes have been documented for animal species in the Late Pleistocene \cite{Baumann2023PaleoSynanthropicNiche}, and, more relevantly for plant communities, in the construction of approximately 4700 artificial forest islands within seasonally flooded Amazonian savannahs during the Early Holocene \cite{Lombardo2020EarlyHoloceneAmazonia}, as well as in the deliberate enrichment of palm species within closed-canopy Amazonian forests through polyculture agriculture and low-severity fire management \cite{iriarte2020106582}. However, the pattern could also reflect preferential settlement at lower elevations where palms were already naturally more abundant due to factors such as soil moisture, slope stability, canopy succession, or other edaphic conditions. Distinguishing between these alternatives requires additional paleoecological data capable of resolving temporal sequences. Although very dense palm clusters occurring near the lower limits of their natural elevation are unlikely to serve as stand-alone predictors of archaeological sites, they provide valuable supplementary information when integrated with broader environmental, multispectral, and hyperspectral datasets. The observed elevational patterns should be understood as contextual rather than diagnostic, but the palm distribution detected in our area of work in the SNSM could be an example of long-term human-ecological interactions and niche construction \cite{laland2000niche, Smith2011GeneralPatternsNicheConstruction}, where sometimes subtle but persistent anthropogenic influences shape the distribution of key plant species in time and space. 

A multi-proxy study by Bush et al. \cite{bush2025ecological} offers complementary evidence for the patterns we observe. They found that modern distributions of \textit{Dictyocaryum} in the northern Andes reflect long-term human influence, showing that contemporary palm locations can carry the signature of past activities. Their approach, using different methods and data sources, favours a combined explanation involving indirect human legacy and climate-driven fluctuations for the increased abundance of \textit{Dictyocaryum}. Their findings support our conclusion that palm clustering can mark areas where people once lived and worked intensively. The fact that independent approaches converge on this interpretation strengthens the case that human activity has shaped modern palm distributions across Andean and neotropical forests. Our spatial, remote-sensing–based findings fit well alongside their multi-proxy evidence, and we believe future studies that explicitly combine both approaches would help test causality more rigorously and assess how generalizable these patterns are across regions.

The current palm distribution is potentially the result of ecological processes that started during the Pre-Columbian occupation of these sites. The composition of the SNSM forest in areas without reported human presence indicates that palms are a key component of the floristic assemblage but their frequency in the vegetation is relatively low \cite{fortier_diversity_2024, Dechner2007}. Although short-term successional dynamics following forest disturbance are relatively well understood for lowland Amazonia \cite{Norden2015}---for instance, the dominance of early successional species depending on disturbance scale---such processes in neotropical montane environments remain less studied \cite{sales_potential_2022,bush2025ecological}. Our data suggest that patterns of certain plant frequencies coincide with nearby archaeological sites. Further research using archaeobotanical proxies---such as phytoliths, environmental DNA extracted from archaeological deposits and surrounding soils, ethnobotanical information, and experimental studies of palm responses to anthropogenically modified soils---is needed to clarify which human behaviours may have contributed to these patterns. The methodology developed here, especially in combination with the referred methods, provides a robust framework for detecting subtle ecological signatures of past human activity and can be applied to other neotropical and tropical environments to further refine our understanding of ancient landscape modification.

\subsection*{Settlement Size and Legacy Strength}

The contrast between the strong palm enrichment associated with Teyuna and the weak or absent signal near numerous much smaller sites suggests that the magnitude of ecological legacy effects may scale with settlement intensity and infrastructure investment. This is consistent with our results showing much higher IDW values around Teyuna than around other centroids. However, our present analyses do not establish a single, universal “threshold” settlement size above which legacies will always be detectable. Legacy strength is influenced by a number of factors, including post-abandonment recovery, local topography, plant management type and intensity, and length of occupation. Nonetheless, the PABAM framework is well suited to address this knowledge gap because it couples species-level detection across continuous landscapes with archaeological location data, enabling comparative analyses of palm density versus site area, infrastructural investment, or occupation length. Future work should formally test for thresholds by relating IDW or cluster metrics to independent estimates of site size, built infrastructure, and dated occupation length, and by applying breakpoint or piecewise regression analyses across larger samples of sites. Such tests will clarify whether there is a minimum settlement footprint or intensity of management that reliably produces long-lasting ecological legacies, or whether legacy strength instead varies continuously with multiple interacting factors.

While we focused our archaeological analysis on the less-disturbed eastern regions to avoid interference from modern activity, we calculated the Spearman correlation across the entire study area to test whether contemporary human presence might explain the palm patterns we observed. This analysis also provides a rough estimate of whether dispersed historic farmsteads or small nucleated settlements may have contributed to higher palm occurrence. The weak correlation ($\rho = 0.21$) indicates that historic settlements, concentrated mainly at lower elevations in the west, do not coincide with the high-density palm clusters found near pre-Columbian infrastructure in the east. This finding demonstrates that contemporary human activity does not explain the observed palm patterns.

\subsection*{Paleoecological Evidence for Anthropogenic Palm Enrichment}

Although Teyuna has been occupied since around 400 CE \cite{Giraldo2010}, paleoecological reconstructions indicated an increase in \textit{Mauritia} and Arecaceae pollen in the Cienaga de Santa Marta, one of the sedimentary sinks draining the eastern flank of the SNSM \cite{vanderhammenNoldus1984}. At Teyuna, \textit{Dictyocaryum lamarckianum} \cite{Paniagua-Zambrana2020} and Arecaceae pollen were detected \cite{Herrera1984Palynological}, coinciding with the first centuries (1100-1250 CE) of the Tairona period, a time of major infrastructural investment in the SNSM. Specifically, pollen data showed a pronounced peak in undifferentiated palm types during this interval, followed by a second, much less pronounced peak around 1650 CE, suggesting that palm abundance increased episodically rather than following a linear long-term trend. However, the age–depth models underlying these reconstructions are approximate, relying on interpolated dates from other cores and excavations, and thus the chronology of these peaks should be interpreted cautiously. Archaeological phytolith assemblages from Teyuna \cite{Giraldo2010} showed the presence of palm morphotypes---primarily spheroids with sinusoidal peripheries and granulated surfaces---dating from around 576 CE onward. These assemblages do not, however, include conical forms associated with \textit{Dictyocaryum} palms, nor do they clearly capture the 1100–1250 CE peak observed in the pollen record. Together, these lines of evidence suggest a potential link between human occupation and increased palm representation. Further investigation using complementary proxies and higher-resolution chronological control is required to clarify the timing and extent of anthropogenic palm enrichment.

\subsection*{Assessing Cluster Stability and Interpreting Cluster Signals}

Clustering outcomes are sensitive to threshold selection, requiring careful parameter optimization and ecological interpretation. Our analysis (Figure \ref{fig:HDBSCAN_parameter_sweep}) shows that low minimum cluster size values generate hundreds to thousands of highly localised clusters, many of which are likely spurious, whereas very large values reduce cluster numbers and obscure localised spatial patterns. By identifying the point at which key metrics stabilize, we selected a minimum cluster size that balances robust cluster detection with conservative noise handling, allowing us to focus on the most spatially coherent and dense palm concentrations while excluding incidental background occurrences lacking clear spatial structure. Importantly, palm clustering is common across the study area and reflects a range of ecological processes independent of human activity. Accordingly, we do not interpret the mere presence of palm clusters as evidence of anthropogenic influence. Instead, our analysis focuses on differences in cluster density and spatial coherence. Clusters that spatially incorporate high-investment archaeological sites---most notably Teyuna---exhibit substantially higher palm densities and larger spatial extents than clusters occurring in areas without archaeological occupation. In contrast, the majority of detected palm clusters are smaller, less dense, and spatially isolated, showing no association with known archaeological features. These clusters likely formed through natural processes and serve as a useful baseline against which the archaeology-linked clusters can be evaluated.

The spatial clustering pattern within Cluster 15 is qualitatively consistent with the aggregated patterns observed across the majority of smaller HDBSCAN clusters in the multi-group analysis. Palm detections are significantly more clustered than expected under CSR across virtually all evaluated spatial scales. However, Cluster 15 is qualitatively distinct from all other clusters in two important respects. First, its spatial extent — approximately 100 km$^2$ containing $\sim$64000 palm detections — far exceeds that of any other cluster, suggesting it represents not a discrete stand or grove but a broad landscape-scale concentration of palms. The co-occurrence of this landscape-scale aggregate with archaeological evidence raises the possibility that Cluster 15 reflects a palimpsest of long-term human presence in the landscape. We tentatively suggest that the exceptional size and density of this cluster may be at least partly attributable to the presence of archaeological sites within its extent, where repeated or sustained human occupation could have promoted palm establishment, dispersal, or maintenance over time. This interpretation remains speculative and would require dedicated ground-truthing and closer integration with the archaeological record to be evaluated further. 

\subsection*{Potential Applications, Limitations, and Future Work Directions} 

Our approach could be extended to analyse other neotropical forests, such as the Chocó–Darién moist forests, the Pacific Equatorial Forest,  the Amazon, and the Eastern flank of the Andes, as well as more arid landscapes with sparse vegetation---where plant species may serve as especially clear ecological markers of past human activity. These markers may reflect lasting ecological legacy of indigenous land use \cite{pavlik2021} or post-abandonment conditions shaped by earlier human presence \cite{Sapircontevegetarchsites2019}. Combining automatic species detection and botanical field surveys with archaeological and paleoenvironmental data can enhance our understanding of pre-Columbian indigenous settlements in the neotropics, and contextualize them within interconnected local and global dynamics of past human–environment relationships \cite{hambrecht2021}. Beyond archaeological applications, the model has potentials for efficient palm tree distribution mapping \cite{FERREIRA2020} for bio-economic development, land-use planning, and ecological monitoring. Our method, using satellite imagery and deep learning, provides a scalable approach to detect such patterns on wider surfaces and more remote regions. By automatically mapping the distribution and dominance of specific plant species---particularly those with cultural or ecological ties to past human groups---this approach can aid in identifying affiliation of current indigenous groups to land, complementing traditional field-based evidence.

Although the palm detection model used in this study successfully captured the overall distribution and density patterns of palm trees throughout the canopy of the study area, it occasionally failed to accurately detect individual trees. These errors are primarily caused by visual similarities between palm trees and certain human-made structures, such as crossroads or buildings, that can exhibit textures or shapes resembling the star-like appearance of a palm tree crown. These misidentifications mainly occur in heavily modified landscapes; in well-preserved forests---where this approach is intended---the risk of confusing natural features with roads or buildings is minimal. To address this issue, future work can focus on integrating semi-supervised learning into the detection framework to enhance palm tree detection performance, even in urbanised landscapes. By combining limited labelled data with large unlabelled image collections, these methods can help models learn more robust and generalizable features. Techniques such as pseudo-labelling \cite{zhou2023ssda} and consistency regularization \cite{zhang2021domain} may be particularly effective in leveraging the variability present in unlabelled data, including differences in appearance, lighting, and environmental conditions. This strategy is expected to improve detection accuracy and adaptability to unseen regions and palm tree varieties.

The computer vision model detected palms with a specific crown pattern visible at the top of the canopy. Palms with different crown patterns or palms whose crowns were located lower in the canopy were unlikely to be detected by our method. Some areas in the study region---particularly in the southwest---show signs of modern human intervention, while the area around Teyuna appears more preserved. To further enhance the efficiency and accuracy of palm detection, future research could focus on integrating active learning into the training pipeline \cite{rodriguez2021mapping}. This would involve iteratively training a model on a small labelled set, using the model to identify the most informative unlabelled samples (e.g., those with high prediction uncertainty), acquiring annotations for these selected samples, and retraining the model with the expanded labelled set. This iterative process has the potential to significantly improve performance while minimizing annotation costs.

Our findings highlight the challenges inherent in attributing modern vegetation patterns to ancient human activities in complex tropical landscapes. Rather than providing definitive evidence of anthropogenic forest modification, our approach offers a scalable, hypothesis-generating framework for identifying candidate areas where long-term human–environment interactions may have been particularly intense. Palm clustering can be interpreted as a supplementary ecological indicator that can help prioritize locations for targeted archaeological, paleoecological, and ecological investigations aimed at clarifying the relative contributions of human actions and natural processes in shaping neotropical montane forests.

The palm tree detection dataset presented here represents a significant contribution to future research across multiple disciplines. In ecology, it provides a well-annotated benchmark for studying the distribution and dynamics of Dictyocaryum sp. in Andean cloud forests, species whose spatial patterning is increasingly understood as a legacy of pre-Columbian land management. Researchers can use this dataset as a labelled source domain for domain adaptation experiments, investigating how models trained on Pleiades Neo imagery can be transferred to comparable montane or lowland tropical environments, substantially reducing annotation burden in new target regions. 

In archaeology and heritage protection, the PABAM framework demonstrated here opens new avenues for non-invasive landscape survey. The integration of deep learning-based vegetation detection with spatial clustering and archaeological site data offers a scalable, reproducible methodology that could be extended to other regions of the Neotropics where ancient human–plant relationships have shaped modern forest composition. Palm assemblages, as proxies for past bio-cultural activity, may help prioritise areas for pedestrian survey, guide heritage impact assessments, and inform protective zoning decisions where surface archaeology is difficult to detect or where sites face pressure from agricultural expansion or infrastructure development. As high-resolution satellite constellations become more accessible, methods like those presented here could be embedded in routine heritage monitoring workflows, enabling conservationists and heritage managers to detect, track, and protect anthropogenic landscapes at regional scales.

\section*{Conclusion}

Our study demonstrates that palm density can serve as an ecological indicator of areas with pre-Columbian settlements in neotropical forests. By applying a scalable methodology that links fine-scale plant distributions with archaeological locations, we document an extensive and dense cluster at the landscape level and natural range distributional shifts towards lower elevations for archaeological site-associated palms. These patterns are consistent with the hypothesis that pre-Columbian populations may have actively managed local environments in ways that favoured palm proliferation. While natural ecological processes cannot be entirely ruled out, the density and extent of palms around settlements with engineered landscape features, and paleoecological evidence suggest that the presence of ancient settlements may have influenced these plant communities over long timescales. Our approach provides a robust framework for detecting subtle ecological legacies of past human activity and highlights the potential of integrating remote sensing, species detection, and archaeological data to uncover long-term human–environment interactions in neotropical montane forests.

These findings contribute to our understanding of ancient human-environment interactions. The integrative approach demonstrated here, combining automated vegetation detection with archaeological data, has particular relevance for tropical regions where traditional archaeological methods face significant challenges due to dense vegetation and limited accessibility. The approach is scalable to any large area with submetric resolution satellite imagery available. This work underscores the importance of interdisciplinary collaboration between ecologists, archaeologists, and computer scientists to uncover the complex legacies of ancient human activities in shaping contemporary forest compositions.

\section*{Materials and Methods}

\subsection*{Pre-Columbian Settlements and Forest Management in the Sierra Nevada de Santa Marta}

Pre-Columbian communities of the Tairona cultural tradition transformed the SNSM landscape. Human occupation, dated from 200 to 1600 CE \cite{Giraldo2010, Rodriguez_et_al_2023}, is divided into two archaeological periods: Neguanje (200–1100 CE) and Tairona (1100–1600 CE). Settlements occur on ridgelines, valley slopes \cite{Rodriguez_et_al_2023,Giraldo2010,serje1987arquitectura}, riverbanks, bays, inlets, flat coastal areas, and deeper canyons \cite{dever2007social,Giraldo2010}. Long-distance mobility between sites within the SNSM has been documented both historically \cite{Gutierrez2022} and in the present, with a 60 km round-trip between two Indigenous communities completed in 18 hours \cite{Yepez2020}. Entire valleys, typically 30 km long, can be traversed in three days from sea level to approximately 3000 m elevation \cite{Reichel1982}. Spanning from coastal lowlands to 2000 m elevation, the Tairona sites feature terraced platforms with curved stone walls, rammed earth foundations, rainwater drainage systems, and circular stone floors for wood-and-thatch structures connected by paved paths \cite{cadavid1985manifestaciones,Giraldo2010,groot1985arqueologia, Herrera1984Agricultural,serje1987arquitectura,Mayorga2024}. All these evidences represent a significant infrastructure investment given the rugged terrain and the logistical challenges of transporting large amount of stones.

Small settlements ($<$3 ha) are scattered across the landscape \cite{Soto2025}. Larger sites ($\geq$3 ha) are less common but present in both the Neguanje and Tairona periods. However, estimating the size of settlements in the Neguanje phase is challenging due to the superimposing Tairona structures that cover the earlier deposits \cite{Giraldo2010}. The largest known settlement is Teyuna (Ciudad Perdida or Buritaca-200), occupied in the Neguanje period from around 400 CE \cite{Giraldo2010}. During the Tairona period, it expanded to a settlement of 33 ha with typical regional architecture \cite{Giraldo2010,Rodriguez_et_al_2023,Mayorga2024}. A remote sensing analysis estimates the footprint of Teyuna at approximately 100 ha (Figure \ref{fig:studyArea}), supporting the role of the settlement as a regional centre within an extensive and interconnected system, but also suggesting a broader anthropogenic landscape surrounding the site. This aligns with the models of “low-density agricultural urbanism” seen in Mesoamerica and Southeast Asia \cite{Rodriguez_et_al_2023} and with evidence from other neotropical forests such as the Amazon \cite{iriarte2020106582}.

Palynological analyses \cite{Herrera1984Agricultural,Herrera1984Palynological} in a radius of 6 km from Teyuna, along with phytolith studies at the site \cite{Giraldo2010}, indicate that landscape transformation began between 600 and 900 CE, likely for small-scale cultivation of maize (\textit{Zea mays}) and manioc (\textit{Manihot} sp.). Phytoliths from \textit{Bactris} sp. and other palms (Arecaceae) were identified in Teyuna, suggesting the exploitation of such plant resource \cite{Giraldo2010}. Also, an increase in palm pollen between 1100 and 1250 CE in the lowlands \cite{vanderhammenNoldus1984}, and at 900-1300 m elevations \cite{Herrera1984Palynological}, coincided with the Tairona occupation period. Among palms, \textit{Dictyocaryum lamarckianum}---common in the SNSM at elevations of 900-2000 m---is used for construction and may represent one of the sources of phytoliths within the recovered assemblages \cite{Giraldo2010, van_der_hammen1986}. Palynological evidence also suggests rapid forest regeneration after site abandonment in the 16th century, implying that Tairona settlements had an ecological footprint on modern forest composition \cite{Herrera1984Agricultural}. Here, “regeneration” refers to reforestation after abandonment, but it does not imply that post-abandonment forests returned to an identical pre-occupation structure; resolving structural differences through time would require additional time-resolved paleoecological proxies beyond the scope of this study.

We selected a study area of approximately 765 km$^2$ in Colombia, known to contain Tairona occupation. We compiled archaeological legacy data (Supplementary Material, Data S1 and S2), harmonising geospatial locations from historical maps and archaeological reports. Harmonising these diverse datasets involved georeferencing historical maps using stable topographic features as ground control points, and extracting legacy coordinate data and textual location descriptions from archaeological reports to project them into a unified spatial reference system (WGS84). We assigned location certainty to these estimated site areas based on the spatial precision of their original sources. Specifically, low and medium certainty sites were identified by analysing and georeferencing these legacy sources, complemented by interviews with local researchers and community members who reported site locations during the 20th century \cite{Mayorga2024}. High-certainty locations were established through systematic pedestrian surveys and by documenting the extents of sites with lidar surveys, as detailed in previous studies \cite{vargas2022arqueologia, Mayorga2024}. Figure \ref{fig:studyArea} presents the study area at various zoom levels for better spatial context. 

Within this region, we used high spatial resolution satellite imagery from the Airbus Pléiades Neo 3/4 constellation acquired between 2022 and 2024 to annotate the locations of star-shaped palm trees. Specifically, we used orthorectified panchromatic (PAN\_ORT) products with radiometric processing to reflectance. The data comprise a single panchromatic band (450–800 nm) with a 30 cm pixel size (nominal ground sampling distance at nadir) and 12-bit radiometric depth. We utilized this panchromatic band because it provides the highest spatial resolution available in the dataset, making it well-suited for detecting individual palm crowns. A total of 47 tiles from 9 Pleiades Neo scenes covering the study area were used for analysis (Supplementary Material, Table S1). A 69.5 km$^2$ non-contiguous subset was annotated with the locations of star-shaped palm trees. To streamline the annotation process and minimize the risk of missing any palm trees, the area was divided into 200x200 m grid (Supplementary Material, Data S4). Human annotators visually inspected each grid and performed heads-up digitization, drawing bounding boxes around each visible palm crown. In total, 24{,}430 palm trees were annotated (Figure \ref{fig:studyArea}, bottom-centre). These annotations (Supplementary Material, Data S3) were used to train and evaluate the YOLO object-detection model, which was subsequently applied to the full imagery to map palms across the study area. Since orbital optical imagery captures reflected radiance from the uppermost visible surfaces (here, the panchromatic band), detections are inherently biased toward crowns exposed at/near the canopy surface; palms located in the understory or obscured by overstory vegetation are less likely to be detected. In the montane forest of the study area, the most likely palm detected under these conditions is \textit{Dictyocaryum} sp. This genus is known for its tall, solitary growth form and prominent canopy presence in Andean cloud forests \cite{Paniagua-Zambrana2020, henderson1995field}. This makes it more readily distinguishable than understory palms such as \textit{Bactris} sp. or lowland species such as \textit{Phytelephas} sp. 

\subsection*{Past Areas of Bio-cultural Activity and Management (PABAM) Methodology} 

We propose a multi-component machine learning method to estimate areas of potential ancient human management using known archaeological sites and vegetation patterns as inputs. In this study, we used star-shaped palm tree crowns as indicators of plants exploited for raw materials in the past and archaeological sites categorized by varying levels of location certainty. 

First, we trained the extra-large variant of the “You Only Look Once” (YOLO) object detection architecture, a family of deep learning models designed for fast and accurate object localisation and classification in images \cite{THU-MIGyolov10}. Within its 10th variant, YOLO introduces several improvements in efficiency and accuracy, and the x configuration represents the most accurate variant. Although satellite imagery is static and does not require real-time inference, we used a one-stage detector (YOLOv10x) to support scalable, wall-to-wall mapping across large areas of 30-cm imagery. Recent advances in the YOLO family—particularly the v10 extra-large configuration—have substantially narrowed the historical accuracy gap with two-stage detectors and perform well on small, visually complex targets such as star-shaped palm crowns. In addition, our workflow involves extensive tiled inference and iterative evaluation of confidence thresholds and post-processing (e.g., clustering and sensitivity checks); a one-stage model provides a simpler, more reproducible pipeline and enables faster iteration than a multi-component two-stage approach. We applied YOLOv10x to 30 cm resolution Pléiades Neo satellite imagery to identify palm trees (Figure \ref{fig:method}), taking advantage of its ability to capture fine-scale canopy features. The labelled palm area was divided into two spatially distinct regions (Figure \ref{fig:method}). The lower region, comprising 47\% of the annotations, was used for training and validation, while the upper region, containing the remaining 53\%, was reserved for testing. This spatial separation aimed to reduce the risk of spatial bias during classification. Each region was further subdivided into 640×640 pixel patches for model input. Within the training and validation region, 95\% of the patches were used for training and the remaining 5\% were used for validation to guide hyperparameter tuning and model selection. The model achieving the highest precision and recall scores on validation data, with hyperparameters detailed in Table S2 (Supplementary Material), was applied to the test patches for evaluation, as well as to the full study area, for palm tree detection  (Supplementary Material, Data S5).

Next, we applied Hierarchical Density-Based Spatial Clustering of Applications with Noise (HDBSCAN) \cite{CampelloHDBSCAN}, which handles clusters of varying densities, does not require a parameter for neighbourhood radius, and is robust to noise and outliers. A critical parameter in HDBSCAN is minimum cluster size, which defines the minimum number of points required to form a cluster. Small values produce numerous small clusters potentially appearing only at extreme density thresholds, whereas larger values merge smaller groups into fewer, more stable clusters, that can potentially hide the variability of the patterns observed. To determine an appropriate minimum cluster size \(m\), we performed a parameter sweep with \(m\) ranging from 5 to 600 in increments of 5. For each value of \(m\), we set \(\text{minimum cluster size} = \text{minimum samples} = m\), used Euclidean distance as the similarity metric, and selected clusters using the excess of mass method with \(\alpha = 1.0\). The resulting configurations were evaluated based on the number of clusters, the fraction of points classified as noise (labelled as \( \ell_i = -1\)), and total cluster stability (Figure \ref{fig:HDBSCAN_parameter_sweep}). The number of clusters \(K\) was defined as the number of unique cluster labels excluding noise, \(K = \left| \{\ell_i \mid \ell_i \neq -1\} \right|\). The noise fraction was defined as the proportion of data points labelled as noise relative to the total number of points, \(\text{Noise Fraction} = \frac{1}{N} \sum_{i=1}^{N} \mathbb{I}(\ell_i = -1)\), where \(N\) is the total number of points, \(\ell_i\) is the cluster label of point \(i\), and \(\mathbb{I}(\cdot)\) is the indicator function, equal to \(1\) if the condition is satisfied and \(0\) otherwise. Total cluster stability measured the persistence of clusters across density thresholds in the hierarchical structure, \( \text{Total Cluster Stability} = \sum_{c=1}^{K} S_c,\) where \(S_c\) is the stability (persistence) of cluster \(c\) returned by HDBSCAN. The final \(m\) value was selected by balancing cluster stability with the removal of trivial clusters, while maintaining an acceptable noise fraction. The selected value of \(m\) was then used to identify spatial clusters of palm trees.

We linked these clusters to past human activity by associating them with the centroids of known archaeological sites and low-location-certainty archaeological zones within the convex hull boundaries of each palm cluster. To compare palm concentrations, we generated centroids for archaeological sites and low-location-certainty archaeological zones. We also generated random centroids in the largest palm cluster and in areas showing clustered palms without reported archaeological sites. Palm concentration differences were measured using an Inverse Distance Weighting (IDW) score, which models the density of nearby palm trees by assigning higher weights to those located closer to each centroid. Palm trees within a 1000 m radius of each centroid were included in the IDW calculations, ensuring that the analysis captured localised patterns while excluding influence from distant and ecologically distinct areas. This radius is a conservative estimate of human mobility based on recently reported walking distances and travel times between Indigenous settlements \cite{Yepez2020,Reichel1982}. The score is calculated as \( G = \sum_i \frac{1}{d_i^w} \), where \( G \) is the IDW score, \( d_i \) is the distance from the selected point to palm tree \( i \) (only if \( d_i \leq 1000 \) m), and \( w \) is a decay factor that gives more weight to closer trees. We use \(w=1\) in our analysis. 
Finally, we used bootstrapping techniques to compare palm concentrations around archaeological centroids with those in control zones---palm areas without known archaeological features. For this, we generated 100 control points, randomly sampled from palm cluster regions without known archaeological features. We also generated an additional set of 100 control points within the largest palm cluster but outside 1 km buffers around the archaeological sites to be able to estimate the extent of potential ancient human management influence within the largest palm cluster. For both control sets and the archaeological centroids within the largest palm cluster, we created 1000 bootstrapped datasets. Each dataset consisted of 17 observations sampled with replacement, matching the number of archaeological centroids in the largest cluster. We chose to focus our analysis in the less-disturbed eastern part of the area of study because the areas in the western portion of the study area have undergone more modern human intervention.

The spatial distribution of palm detections inside each cluster was characterised using Ripley's K-function and its linearised transformation, the L-function. Two independent analytical protocols were applied: a multi-group analysis covering all HDBSCAN clusters (except Cluster 15) and a dedicated analysis of Cluster 15, which required a distinct methodological treatment due to its exceptional spatial extent.

For the Ripley's K analyses palm detection coordinates were projected from geographic coordinates to UTM Zone 18N  to ensure all distance calculations were performed in metres. The study window for each group was defined as an alpha-shape fitted to the sampled point pattern ($\alpha$ = 0.005). The maximum search radius (r\_max) was set to one quarter of the shortest bounding box dimension of the study window, subject to a cap of 2000 m for windows exceeding 50 km$^2$ (if any). Two categories based on archaeological association were analysed independently: clusters spatially coinciding with known archaeological sites (Group A, n = 2: clusters 1 and 2) and clusters with no such association (Group B, n = 18: clusters 3–14 and 16–21). Differences in clustering metrics between both groups were tested using two-sided Mann-Whitney U tests, with effect size reported as the rank-biserial correlation r. Statistical analyses were conducted at $\alpha$ = 0.05. 

Cluster 15 was analysed separately due to its exceptional size, comprising approximately 64000 palm detections distributed across roughly a convex hull of 100 km². Applying the simple random subsampling and fixed window approach used in the multi-group analysis would have yielded a point density far too low relative to the window area for meaningful spatial statistics: the CSR null model would have been evaluated against a sparsely sampled representation of a large landscape rather than against the local-scale spatial structure of ecological interest. A dedicated protocol was therefore developed. 

Prior to subsampling, the full extent of Cluster 15 was divided into a regular grid of 2×2 km cells. Points were then sampled from each occupied cell in proportion to its local point density, so that the subsample preserved the large-scale spatial distribution of detections across the cluster. The total subsample size was set to 5\% of the full cluster count, up to a maximum of 10000 points, generating a substantially higher sampling fraction (n=3202) than the total detected palms of other clusters. The study window was defined as an alpha-shape fitted to the stratified subsample ($\alpha$ = 0.005). The maximum search radius was fixed at r\_max = 500 m. This choice was deliberate to assess clustering at fine spatial scales consistent with the ecological processes driving palm aggregation, rather than at landscape scales where the large window area would dilute the clustering signal. 

The corrected Ripley's K and centred L-function were computed identically to the multi-group analysis. A 95\% pointwise simulation envelope was constructed from 199 Monte Carlo CSR realisations.  Simulations were seeded by spawning 199 independent child seeds from a single root seed (42) using np.random.SeedSequence, ensuring both reproducibility and statistical independence between simulations. Given the differences in sampling design, window area, and r\_max scale, the Cluster 15 results are reported as a standalone characterisation of within-cluster spatial structure and are not included in the comparative statistical tests between archaeological groups. 

For both protocols, Ripley's K\-function with isotropic edge correction was computed following Ripley \cite{ripley1976second,ripley1988statistical}, and the centred Besag’s  L-function transformation was applied \cite{besag1977contribution}. The L-function was evaluated at 50 equally spaced distance values from 0 to r\_max. Summary metrics extracted from each L-function curve included the maximum observed value of L(r) $-$ r across all evaluated distances (L\_max), reflecting the peak clustering intensity;  the distance at which this maximum occurs (r\_at\_L\_max), indicating the spatial scale at which clustering is strongest; the integral of L(r) $-$ r over the full range of evaluated distances (L\_integral), providing a single measure of overall clustering intensity; the smallest (r\_sig\_first) and largest (r\_sig\_last) distances at which the observed curve significantly exceeded the upper CSR envelope, defining the range of spatial scales over which clustering was statistically significant;  and the percentage of evaluated distance bins where the observed L(r) $-$ r exceeded the upper envelope (pct\_r\_sig), quantifying the breadth of the significant clustering signal relative to the total range of distances examined.

We further evaluated the consistency of the palm detection model by comparing the elevation of detected palms with the elevation distribution of \textit{Dictyocaryum} occurrences recorded in the Global Biodiversity Information Facility (GBIF) database \cite{gbif_Dictyocaryum}. Elevation data for both sets were extracted from the Shuttle Radar Topography Mission (SRTM) dataset at 1 arc second (approximately 30 m) resolution \cite{NASA2013}. An interquartile range-based outlier filtering on the SRTM elevation was applied to the GBIF baseline. GBIF occurrence records were used as an observation-based reference for the species’ reported elevational envelope; however, because GBIF data can reflect uneven sampling effort and potential anthropogenic influences, we interpret this comparison as descriptive rather than as a definitive estimate of a pristine natural range. We then compared GBIF data with the elevations of palm trees detected by our deep learning model within a 1000 m radius of archaeological centroids in the largest palm cluster. We also examined the elevation data for all palm trees located in areas showing clustered and non-clustered palms without archaeological sites. 

To assess the potential relationship between the number of modern human activities and the number of palm trees detected within the study area, we conducted a Spearman rank correlation between the number of modern buildings detected in a 200x200 m cell grid and the detected palms (Figure \ref{fig:footprint_and_palms}B). The analysis used a dataset containing automatic detected building from satellite imagery between 2014 and 2024 \cite{microsoft2023global} and palm trees detected per spatial unit (Figure \ref{fig:footprint_and_palms}B).

\subsection*{Software and Computational Tools}

Data processing and statistical analyses were performed using Python (Version 3.11.12; Python Software Foundation; https://www.python.org), employing the libraries NumPy, SciPy, Pandas, GeoPandas, Shapely, Matplotlib, Seaborn, alphashape, joblib, and HDBSCAN. Geographic data cleaning, validation, and visualization were conducted using ArcGIS Pro (Version 3.2.3; Environmental Systems Research Institute, Redlands, CA, USA; https://www.esri.com/en-us/arcgis/products/arcgis-pro). To train the YOLOv10 detector, we used Ultralytics YOLO (ultralytics v8.3.160; https://pypi\\.org/project/ultralytics/), a Python package built on the PyTorch deep-learning framework (https://\\pytorch.org/).

\clearpage 

%
\bibliography{Ref} 
\bibliographystyle{sciencemag}

%
%
%
%
%
%


\section*{Acknowledgments}
\paragraph*{Funding:}
This research is part of the "Mapping the Archaeological Pre-Columbian Heritage in South America – MAPHSA" Project funded by Arcadia – a charitable fund of Lisbet Rausing and Peter Baldwin. This research was also supported by a Leiden-Delft-Erasmus Global Support Grant (Grant No. LDE-127). Pleiades Neo satellite imagery data provided by the European Space Agency.
\paragraph*{Author contributions:}
Conceptualization: S.F., M.Ma., S.M., J.G.S; Data curation: S.F., M.I.M., F.M., K.P., C.R., J.C.V.; Formal analysis: S.F., S.M.; Funding acquisition: S.F., M.Ma., J.G.S.; Investigation: S.F., S.M.; Methodology: S.F., S.M.; Project administration: S.F., M.Ma.,M.M., J.G.S., J.C.V., F.W.T.; Resources: C.A., F.M., F.W.T.; Supervision: S.F., M.Ma., F.W.T.; Visualization: S.F., S.M.; Writing – original draft: S.F., S.M.; Writing – review \& editing: All authors
\paragraph*{Competing interests:}
There are no competing interests to declare
\paragraph*{Data and materials availability:}
All data supporting the findings of this study are either cited in the references or included in the supplementary materials.

\subsection*{Supplementary materials}
Supplementary material file\\
Data S1\\
Data S2\\
Data S3\\
Data S4\\
Data S5\\


\newpage


\renewcommand{\thefigure}{S\arabic{figure}}
\renewcommand{\thetable}{S\arabic{table}}
\renewcommand{\theequation}{S\arabic{equation}}
\renewcommand{\thepage}{S\arabic{page}}
\setcounter{figure}{0}
\setcounter{table}{0}
\setcounter{equation}{0}
\setcounter{page}{1} 


\begin{center}
\section*{Supplementary Materials for\\ \scititle}

    Sebastian Fajardo$^{1\ast\dagger}$,
    Sina Mohammadi$^{1\ast\dagger}$,

    Jonas Gregorio de Souza$^{2}$,
    César Ardila$^{3}$,\\
    Alan Tapscott$^{4}$,
    Shaddai Heidgen$^{4}$,
    Maria Isabel Mayorga Hernández$^{5}$, \\
    Sylvia Mota de Oliveira$^{6}$,
    Fernando Montejo$^{3}$,
    Marco Moderato$^{4}$, \\
    Vinicius Peripato$^{7}$,
    Katy Puche$^{3}$,
    Carlos Reina$^{3}$, \\
    Juan Carlos Vargas$^{8}$,
    Frank W. Takes$^{1}$,
    Marco Madella$^{4,9,10\ast}$ \\
    \small$^\ast$Corresponding authors. Sebastian Fajardo: s.d.fajardo.bernal@liacs.leidenuniv.nl;\\ \small Sina Mohammadi: s.mohammadi@liacs.leidenuniv.nl;\\ \small Marco Madella: marco.madella@upf.edu\\
	\small$^\dagger$These authors contributed equally to this work.

\end{center}

\subsection*{This PDF file includes:}

\paragraph{Captions for Other Supplementary Materials for this Manuscript.}
Explanatory captions for Data S1 to S5.

\paragraph{Scene IDs, Acquisition Dates, and Tiles Used.}
The scene IDs and acquisition dates/times for all scenes are reported in Table \ref{tab:pneo_scenes}  (tile identifiers encoded as R\#C\# in the filenames).

\paragraph{Training Configuration and Hyperparameters for the YOLOv10x Palm Crown Detector.} Training configuration and key hyperparameters for the YOLOv10x are listed in Table \ref{tab:yolov10x_hparams}.

\paragraph{Ripley’s L-function Summary Metrics (n=21) and Mann–Whitney U Test Results.} 
Ripley’s L-function summary metrics (n=21) and Mann–Whitney U test results are reported in Table \ref{tab:Ripley} and \ref{tab:Mann-Wh} respectively.

\paragraph{Ripley’s L-function (L(r) $-$ r) Plots for the Detected Palm Clusters.} 
Ripley’s L-function plots are shown in Figure S1.

\newpage

\subsubsection*{Captions for Other Supplementary Materials for this Manuscript.}


\paragraph{Caption for Data S1.}
\textbf{Archaeological\_Centroids.xlsx}
This dataset contains the approximate central location of reported archaeological sites or zones inside the area of study. The dataset includes the ID\_site field, which provides a unique identifier or commonly used name for each site, and the ID\_source\_field, referencing the source from which the site's ID information was derived. Additionally, the Location\_source field indicates the main source of the geolocation. When this field is marked as $N/A$, it denotes that the coordinates were established as part of the present study. This dataset is provided as GeoJSON Lines (geojsonl) and uses the WGS84 coordinate reference system. The file is provided as a .xlsx, where each row contains a GeoJSON LineString stored as text in a cell. To use the data in GIS software, these GeoJSON entries need to be extracted and saved as a .geojson file.

\paragraph{Caption for Data S2.}
\textbf{Study\_Area.xlsx}
This data includes the area of study. The polygon includes the perimeter in kilometres and the area in square kilometres. It is provided as a GeoJSON Line (geojsonl) and uses the WGS84 coordinate reference system. The file is provided as a .xlsx, where the first row contains a GeoJSON LineString stored as text in a cell. To use the data in GIS software, this GeoJSON entry need to be extracted and saved as a .geojson file.

\paragraph{Caption for Data S3.}
\textbf{Manual\_Labels\_Palms.xlsx}
This  dataset includes manually annotated polygon features representing individual palms. Each polygon is assigned a unique ID. The dataset is provided as  GeoJSON Lines (geojsonl) and uses the WGS84 coordinate reference system. The file is provided as a .xlsx, where each row contains a GeoJSON LineString stored as text in a cell. To use the data in GIS software, these GeoJSON entries need to be extracted and saved as a .geojson file.

\paragraph{Caption for Data S4.}
\textbf{Labeled\_Areas.xlsx}
This dataset includes the area where palms were manually annotated. It is divided in cells of 200 x 200 m. Each polygon is assigned a unique ID, with their area in square metres and their perimeter in metres. The dataset is provided as GeoJSON Lines (geojsonl) and uses the WGS84 coordinate reference system. The file is provided as a .xlsx, where each row contains a GeoJSON LineString stored as text in a cell. To use the data in GIS software, these GeoJSON entries need to be extracted and saved as a .geojson file.

\paragraph{Caption for Data S5.}
\textbf{Detected\_Palms\_04\_Threshold.xlsx}
This  dataset consists of polygon features representing automatically detected palm bounding boxes with a probability $\ge$ 0.4 provided by our deep learning model. Each polygon is associated with a unique identifier. This dataset is provided as GeoJSON Lines (geojsonl) and uses the WGS84 coordinate reference system. The file is provided as a .xlsx, where each row contains a GeoJSON LineString stored as text in a cell. To use the data in GIS software, these GeoJSON entries need to be extracted and saved as a .geojson file.

\newpage
\subsubsection*{Scene IDs, Acquisition Dates, and Tiles Used}
We used 9 Pléiades Neo scenes (acquisitions/products). Because the delivered ortho PAN products are provided as tiled files, these scenes correspond to 47 tiles in total.

\begin{table*}[h]
\caption{Pléiades Neo scenes used in this study and their associated tiles. Acquisition date/time is taken from the DIMAP filename (format \texttt{YYYYMMDDHHMMSSs}; the last digit denotes tenths of a second). Tiles (e.g., \texttt{R1C2}) are parts of a scene/product, not separate scenes.}
\centering
\footnotesize
\resizebox{\textwidth}{!}{%
\begin{tabular}{@{}llllll@{}}
\toprule
Scene ID (from filename) & Satellite & Acquisition date/time & Job ID & \# tiles & Tile IDs used \\
\midrule
\texttt{PNEO3\_202205291533133\_PAN\_ORT} & PNEO3 & 2022-05-29 15:33:13.3 & \texttt{PWOI\_000276922\_1\_1\_F\_1} & 4 & \texttt{R1C1, R1C2, R2C1, R2C2} \\
\texttt{PNEO3\_202303111532587\_PAN\_ORT} & PNEO3 & 2023-03-11 15:32:58.7 & \texttt{PWOI\_000276922\_2\_1\_F\_1} & 4 & \texttt{R1C1, R1C2, R2C1, R2C2} \\
\texttt{PNEO4\_202401301533517\_PAN\_ORT} & PNEO4 & 2024-01-30 15:33:51.7 & \texttt{PWOI\_000276922\_3\_1\_F\_1} & 6 & \texttt{R1C1, R1C2, R2C1, R2C2, R3C1, R3C2} \\
\texttt{PNEO4\_202312091532545\_PAN\_ORT} & PNEO4 & 2023-12-09 15:32:54.5 & \texttt{PWOI\_000276922\_4\_1\_F\_1} & 4 & \texttt{R1C1, R1C2, R2C1, R2C2} \\
\texttt{PNEO3\_202401071540320\_PAN\_ORT} & PNEO3 & 2024-01-07 15:40:32.0 & \texttt{PWOI\_000276809\_5\_2\_F\_1} & 8 & \texttt{R1C1, R1C2, R2C1, R2C2, R3C1, R3C2, R4C1, R4C2} \\
\texttt{PNEO4\_202312091532518\_PAN\_ORT} & PNEO4 & 2023-12-09 15:32:51.8 & \texttt{PWOI\_000276809\_6\_1\_F\_1} & 4 & \texttt{R1C1, R1C2, R2C1, R2C2} \\
\texttt{PNEO4\_202401301533501\_PAN\_ORT} & PNEO4 & 2024-01-30 15:33:50.1 & \texttt{PWOI\_000276809\_7\_1\_F\_1} & 6 & \texttt{R1C1, R1C2, R2C1, R2C2, R3C1, R3C2} \\
\texttt{PNEO4\_202305051540322\_PAN\_ORT} & PNEO4 & 2023-05-05 15:40:32.2 & \texttt{PWOI\_000276922\_8\_1\_F\_1} & 3 & \texttt{R1C1, R2C1, R3C1} \\
\texttt{PNEO3\_202403091533140\_PAN\_ORT} & PNEO3 & 2024-03-09 15:33:14.0 & \texttt{PWOI\_000276922\_9\_1\_F\_1} & 8 & \texttt{R1C1, R1C2, R2C1, R2C2, R3C1, R3C2, R4C1, R4C2} \\
\bottomrule
\end{tabular}%
}

\label{tab:pneo_scenes}
\end{table*}

\newpage
\subsubsection*{Training Configuration and Hyperparameters for the YOLOv10x Palm Crown Detector}


\begin{table}[h]
\centering
\caption{Training configuration and key hyperparameters for the YOLOv10x palm-crown detector.}
\label{tab:yolov10x_hparams}
\scriptsize 
\renewcommand{\arraystretch}{1.05}
\setlength{\tabcolsep}{5pt}
\begin{tabular}{p{0.30\linewidth} p{0.64\linewidth}}
\hline
\textbf{Category} & \textbf{Details} \\
\hline
Architecture / initialization & YOLOv10x (pretrained weights enabled) \\
Input / training length & 640$\times$640 patches, 250 epochs, early stopping patience = 100 \\
Batch size \& compute & batch 32, mixed precision (AMP) \\
Optimization & lr0 0.01, final LR factor (lrf) 0.01, momentum 0.937, weight decay 0.0005, warmup 3 epochs \\
Regularization & dropout 0.1 \\
Key augmentations & rotation $\pm$15$^\circ$, translation 0.1, scale 0.5, vflip 0.5, hflip 0.5, mosaic 1.0, random erasing 0.15, RandAugment \\
Post-processing (train/val) & IoU threshold 0.7, max detections 300 \\
\hline
\end{tabular}
\end{table}

\newpage
\subsubsection*{Ripley’s L-function Summary Metrics (n=21) and Mann–Whitney U Test Results.}



\begin{table}[h]
\centering
\scriptsize
\setlength{\tabcolsep}{3pt}
\renewcommand{\arraystretch}{0.95}
\caption{Ripley's L-function summary metrics for each detected palm clusters (n=21). L\_max: maximum observed $L(r)-r$ value; r\_at\_L\_max: distance (m) at which L\_max occurs; L\_integral: area under the $L(r)-r$ curve; r\_sig\_first and r\_sig\_last: first and last distances (m) at which the observed $L(r)-r$ exceeds the 95\% CSR envelope; pct\_r\_sig: percentage of r values showing significant clustering; n\_points: number of palm detections (for Cluster 15 is a sample 5\% of the full cluster count); window area estimated using alpha shape ($\alpha=0.005$).}
\label{tab:XA}
\begin{tabular}{r r r r r r r r >{\raggedleft\arraybackslash}p{1.55cm}}
\toprule
\textbf{CLUSTER\_ID} &
\textbf{L\_max} &
\textbf{r\_at\_L\_max} &
\textbf{L\_integral} &
\textbf{r\_sig\_first} &
\textbf{r\_sig\_last} &
\textbf{pct\_r\_sig} &
\textbf{n\_points} &
\makecell{\textbf{Alpha shape}\\\makecell{\textbf{window area}\\\textbf{(km$^2$)}}} \\
\midrule
1  & 0    & 0    & -148241.14 & N/A   & N/A   & 0  & 138  & 1.1 \\
2  & 50.0 & 149.5& 11781.0    & 20.4 & 312.7& 88 & 130  & 1.1 \\
3  & 47.3 & 213.2& 11210.4    & 12.2 & 298.5& 96 & 192  & 0.5 \\
4  & 20.6 & 95.0 & 3155.6     & N/A   & N/A   & 0  & 444  & 0.8 \\
5  & 54.2 & 180.9& 21871.0    & 12.1 & 591.1& 98 & 789  & 2.8 \\
6  & 24.7 & 111.9& 3788.5     & 12.9 & 210.8& 94 & 103  & 0.3 \\
7  & 33.5 & 80.8 & 13737.9    & 16.2 & 791.9& 74 & 633  & 3.2 \\
8  & 34.9 & 90.6 & 4680.3     & 12.3 & 201.7& 94 & 460  & 1.2 \\
9  & 46.5 & 125.6& 9741.7     & 14.8 & 332.4& 88 & 594  & 1.1 \\
10 & 38.4 & 57.2 & 4784.1     & 13.2 & 202.6& 88 & 180  & 0.6 \\
11 & 62.9 & 105.8& 6822.5     & 12.5 & 152.5& 92 & 117  & 0.1 \\
12 & 31.4 & 95.0 & 6050.5     & 12.7 & 304.2& 94 & 547  & 0.5 \\
13 & 45.3 & 223.8& 10266.6    & 11.8 & 288.5& 96 & 810  & 1.6 \\
14 & 21.5 & 248.2& 3757.9     & 15.2 & 248.2& 94 & 755  & 0.6 \\
16 & 49.3 & 135.6& 11336.7    & 12.3 & 302.0& 96 & 396  & 1.0 \\
17 & 34.1 & 104.1& 5018.1     & 13.0 & 212.5& 94 & 372  & 0.7 \\
18 & 24.0 & 56.8 & 3130.8     & 11.4 & 185.5& 94 & 221  & 0.4 \\
19 & 5.6  & 80.2 & 210.7      & 12.0 & 98.2 & 78 & 493  & 0.1 \\
20 & 3.7  & 25.4 & -41.3      & 11.9 & 40.7 & 22 & 118  & 0.0 \\
21 & 6.2  & 80.9 & 328.9      & 12.3 & 120.1& 84 & 682  & 0.2 \\
15 & 31.3 & 500.0& 11174.2    & 20.4 & 500  & 96 & 3202 & 54.0 \\
\bottomrule
\end{tabular}
\label{tab:Ripley}
\end{table}

\newpage
\begin{center}
\footnotesize
\setlength{\tabcolsep}{5pt}
\renewcommand{\arraystretch}{1.05}
\captionof{table}{Mann-Whitney U test results comparing Ripley's L-function summary metrics between clusters with archaeology (Group A, n=2) and without archaeology (Group B, n=18).}
\label{tab:X2}
\begin{tabular}{l r r r r l}
\toprule
\textbf{Metric} &
\textbf{Median group A} &
\textbf{Median group B} &
\textbf{p-value} &
\textbf{r} &
\textbf{Significance} \\
\midrule
L\_max              & 25        & 33.81   & 0.8526 & 0.111  & not significant \\
r\_at\_L\_max       & 74.77     & 99.57   & 0.6737 & 0.222  & not significant \\
L\_integral         & -68230.09 & 4901.09 & 0.8526 & 0.111  & not significant \\
pct\_r\_sig         & 44        & 94      & 0.1587 & 0.639  & not significant \\
pct\_above\_CSR\_hi & 44        & 94      & 0.1587 & 0.639  & not significant \\
pct\_below\_CSR\_lo & 49        & 2       & 1      & -0.028 & not significant \\
pct\_below\_zero    & 50        & 4       & 0.6496 & -0.222 & not significant \\
mean\_L\_obs        & -70.52    & 20.43   & 0.5895 & 0.278  & not significant \\
\bottomrule
\end{tabular}
\label{tab:Mann-Wh}
\end{center}

\clearpage

\includepdf[
  pages=1,
  scale=0.8,
  pagecommand={
    \thispagestyle{plain}
    \subsubsection*{Ripley’s L-function (L(r) $-$ r) Plots for the Detected Palm Clusters.}
    \label{fig:LfunctionS1}
  }
]{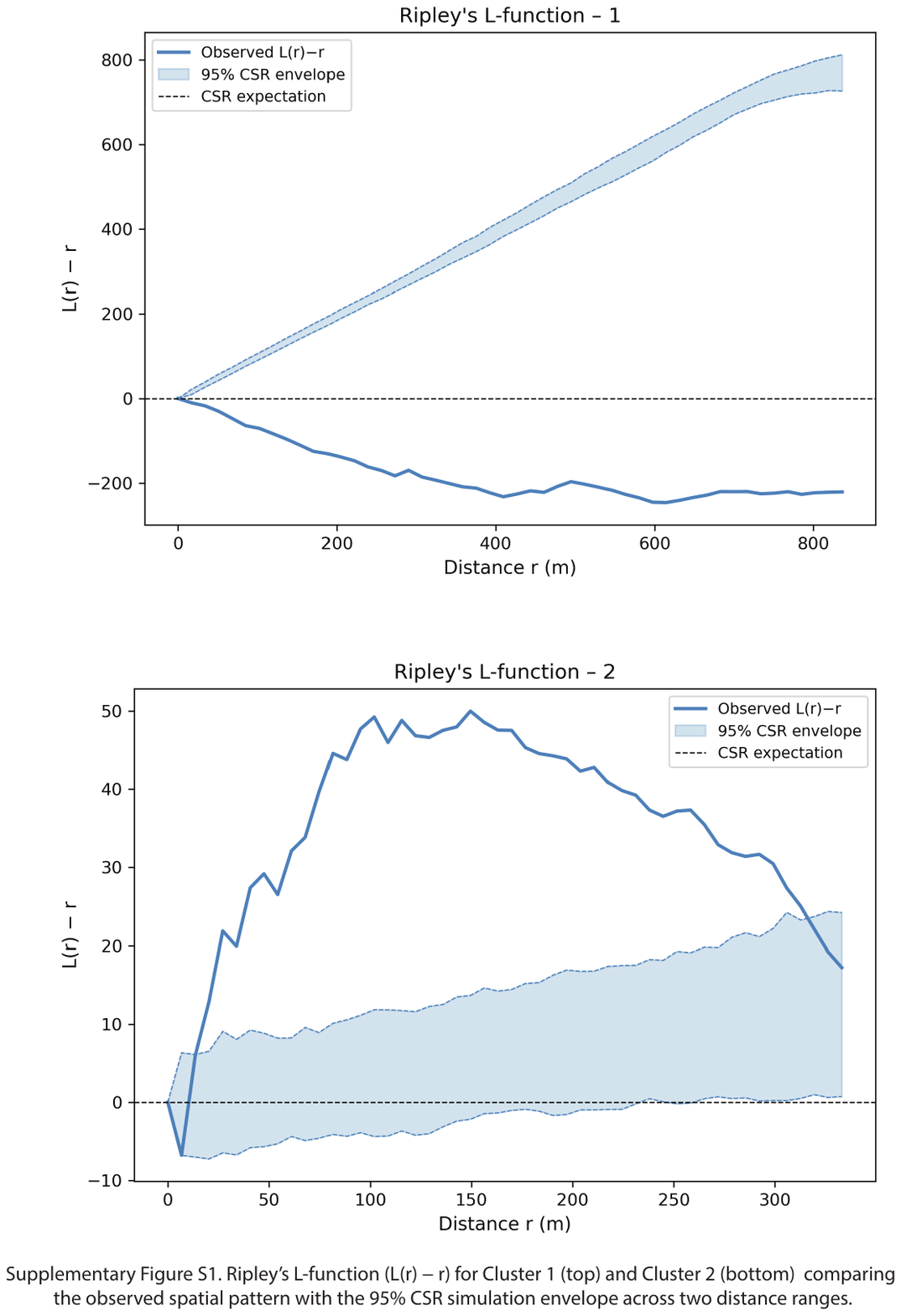}

\includepdf[
  pages=2-,
  scale=0.85,
  pagecommand={\thispagestyle{plain}}
]{Lfunction_clusters1to21.pdf}

\clearpage 





\end{document}